\documentclass[10pt,twocolumn,letterpaper]{article}

\usepackage{iccv}
\usepackage{times}
\usepackage{epsfig}
\usepackage{graphicx}
\usepackage{amsmath}
\usepackage{amssymb}
\usepackage{booktabs}
\usepackage{cancel}
\usepackage{floatrow}
\newfloatcommand{capbtabbox}{table}[][\FBwidth]
\usepackage{multirow}
\usepackage{xcolor}
\usepackage{lipsum}

\usepackage[pagebackref=true,breaklinks=true,letterpaper=true,colorlinks,bookmarks=false,citecolor=mycitecolor,linkcolor=mylinkcolor,urlcolor=myurlcolor]{hyperref}
\definecolor{mycitecolor}{RGB}{0,180,150}
\definecolor{mylinkcolor}{RGB}{200,50,50}
\definecolor{myurlcolor}{RGB}{200,50,200}
\usepackage{cleveref}
\usepackage[accsupp]{axessibility}

\iccvfinalcopy 

\def\iccvPaperID{5275} 

\ificcvfinal\fi

\newcommand*{\affaddr}[1]{#1} 

\newcommand*{\email}[1]{\small \texttt{#1}}

\begin{document}

\title{Learning from Semantic Alignment between Unpaired Multiviews for\\ Egocentric Video Recognition}

\author{%
Qitong Wang$^1$, Long Zhao$^2$, Liangzhe Yuan$^2$, Ting Liu$^2$, Xi Peng$^1$\\
\affaddr{$^1$University of Delaware},
\affaddr{$^2$Google Research}\\
\email{\{wqtwjt,xipeng\}@udel.edu} \hspace{0.5cm}
\email{\{longzh,lzyuan,liuti\}@google.com}\\
}

\maketitle
\ificcvfinal\thispagestyle{empty}\fi

\newcommand\blfootnote[1]{%
\begingroup
\renewcommand\thefootnote{}\footnote{#1}%
\addtocounter{footnote}{-1}%
\endgroup
}

\begin{abstract}
We are concerned with a challenging scenario in unpaired multiview video learning.
In this case, the model aims to learn comprehensive multiview representations while the cross-view semantic information exhibits variations.
We propose Semantics-based Unpaired Multiview Learning (SUM-L) to tackle this unpaired multiview learning problem. 
The key idea is to build cross-view pseudo-pairs and do view-invariant alignment by leveraging the semantic information of videos.
To facilitate the data efficiency of multiview learning, we further perform video-text alignment for first-person and third-person videos, to fully leverage the semantic knowledge to improve video representations. 
Extensive experiments on multiple benchmark datasets verify the effectiveness of our framework.
Our method also outperforms multiple existing view-alignment methods, under the more challenging scenario than typical paired or unpaired multimodal or multiview learning. Our code is available at \url{https://github.com/wqtwjt1996/SUM-L}.
\end{abstract}
\section{Introduction}

Recent years have witnessed a tremendous increase in the research regarding egocentric video understanding and learning~\cite{ego-exo,lu2013story,lee2015predicting,su2016detecting,matsuo2014attention,ego-topo,huang2018predicting,ego-aco,ego-vqa,qi2021self,wang2020symbiotic,furnari2019would}.
In parallel, a recent study~\cite{ego-exo} verifies that the activity of third-person videos in fact positively inform first-person video learning.
Early works~\cite{charades-ego,yu2019see} of egocentric video learning and understanding pay a lot of attention to paired multiview data representation learning.
However, synchronized first-person and third-person video pairs are difficult to collect.
This severely limits the data scale and scope of multiview learning for egocentric data.

Recently, egocentric video datasets~\cite{epic-55, epic-100, ego4d} with larger scale have become available.
However, since head-mounted cameras were not available until recent years, very few first-person videos have paired third-person videos~\cite{lemma}.
Fortunately, there exist third-person videos from different datasets~\cite{ava,kinetics-400}.
It is worth noting that unpaired third-person videos are relatively easier to capture and obtain compared with paired ones.
Therefore, in this paper, we study an interesting yet seldom investigated problem: how to leverage unpaired third-person videos to help egocentric video learning?

\begin{figure}[t]
\centering
\includegraphics[width=1.02\textwidth]{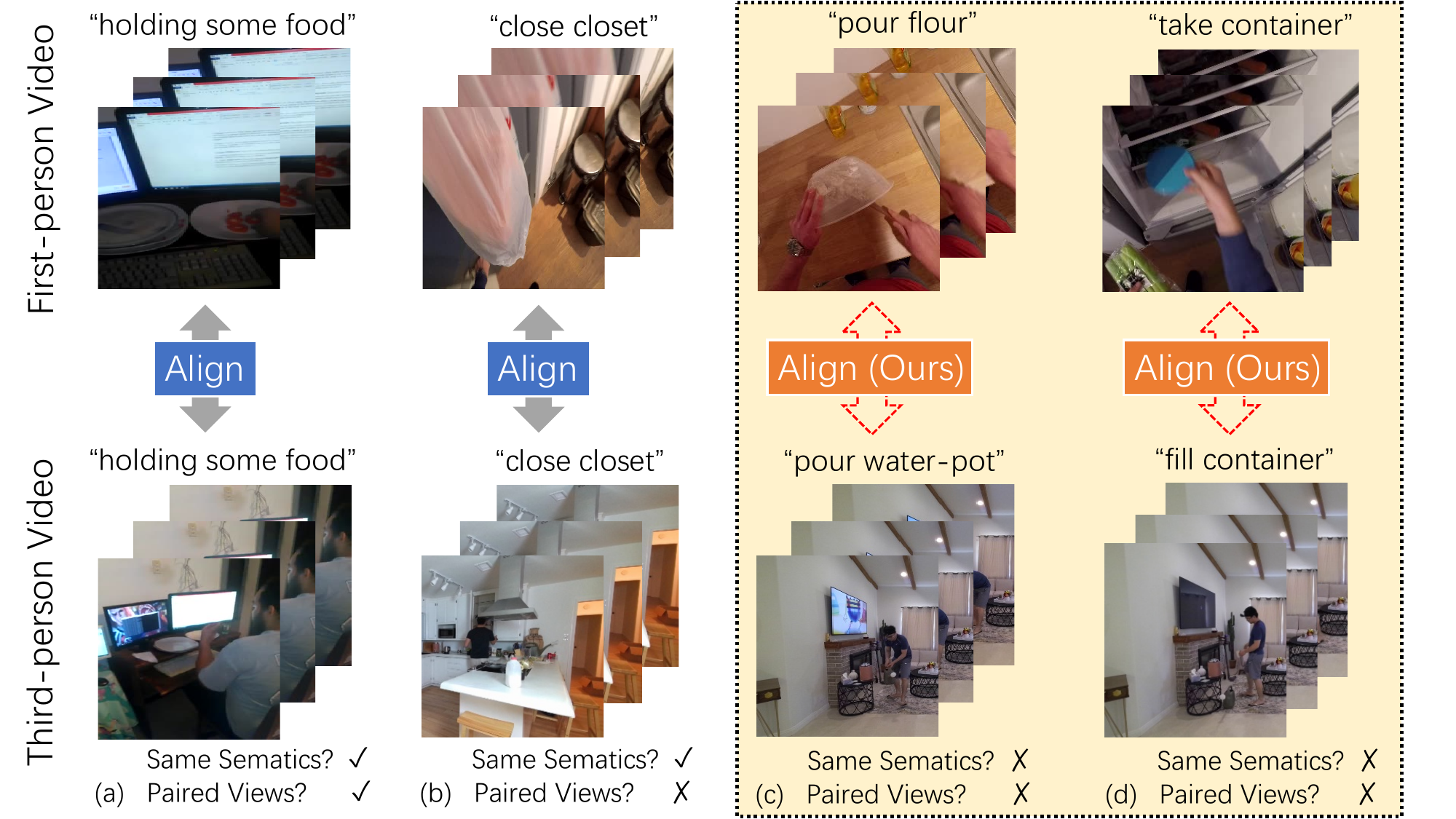}
\caption{(a) The typical multiview video learning method learns view-invariant representations from paired first-person and third-person videos. 
(b) Recent multimodal study~\cite{NEURIPS2021_unpaired_view} is interested in the alignment between unpaired samples which share identical semantics.
In this paper, we study a more challenging scenario than existing works: unpaired multiview samples alignment with partially similar semantics, such as videos with (c) the same verb only, or (d) the same object only.
The verb-object phrases above the video clips denote the semantic information of videos.} 
\label{overview_fig}
\end{figure}

\begin{figure*}[t]
\centering
\includegraphics[width=1.0\textwidth]{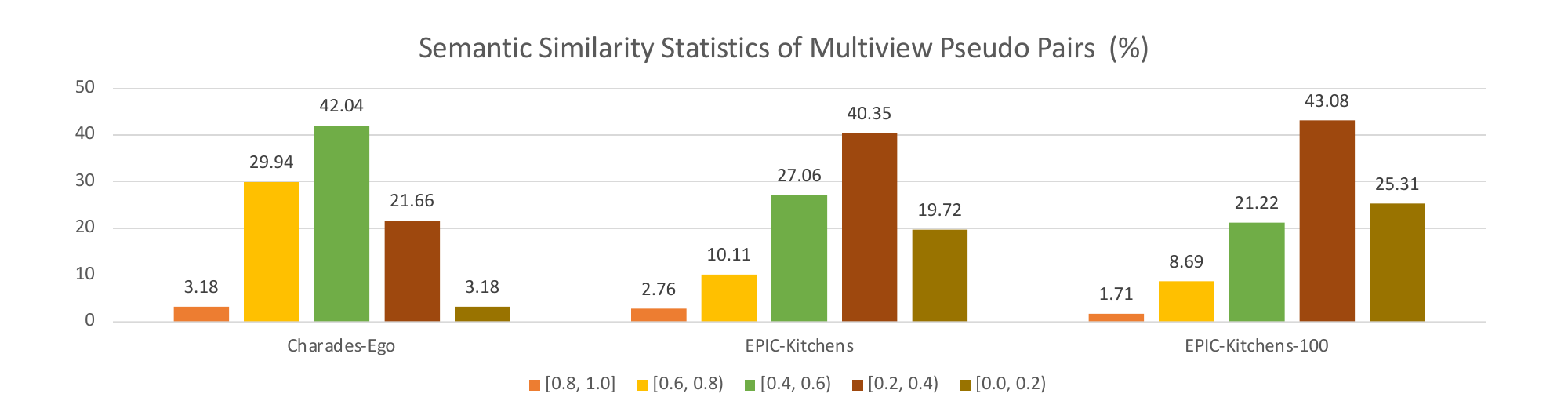}
\caption{Statistics of semantic similarities between the first-person samples in~\cite{charades-ego, epic-55, epic-100} and the most similar single-agent third-person samples they can find from the LEMMA~\cite{lemma} dataset.
The semantic similarity is the cosine distance between textual phrase vectors of videos encoded by a large language model~\cite{text_encoder}.
Each bar in the figure represents the percentage of action classes that fall within one semantic similarity range (\eg, [0.8, 1.0]) in an egocentric dataset compared to all action classes in this dataset.
In all of the three popular egocentric datasets, very few first-person videos are able to find third-person videos with identical semantics.
For reference, in the studies for paired multiview data~\cite{cmc, charades-ego, yu2019see, cv_mim}, all the semantic similarities of cross-view data pairs are (or very close to) ``1.0''. } 
\label{ps_stat}
\end{figure*}

Unpaired multiview learning is rarely studied.
While recent works~\cite{ma2021smil, Ma_2022_tran, shi_mm, Zhao_2020_CVPR} study the scenario where paired modalities are missing.
But they do not leverage the potential of unpaired data from diverse datasets.
Moreover, to align the unpaired multimodal samples, Kundu~\etal~\cite{NEURIPS2021_unpaired_view} propose one relation distillation method to align the unpaired samples.
They assume that unpaired data need to share identical semantic information. 
However, in the unpaired first-person and third-person video learning scenario, for the vast majority of egocentric videos, it is almost impossible to mine the third-person samples with identical semantics.
The only alternatives are \textit{partially semantics-similar} unpaired third-person videos.
For example, for one egocentric video clip with ``cut watermelon" semantics, the most similar third-person video is those with ``cut lemon" (same verb) or ``eat watermelon" (same object) semantics.
Examples and comparisons between our scenario and typical paired or unpaired multiview or multimodal learning are shown in Figure~\ref{overview_fig}, and we display comprehensive semantics similarity statistics of multiview data studied in this paper (see Figure~\ref {ps_stat}).
Therefore, how to employ the unpaired and partially semantics-similar third-person views to help first-person view learning is an open problem.
Intuitively, compared with previous research, our setting is more challenging since most of the first-person videos only share partially similar semantics with third-person views.


In this paper, we are interested in view-invariant alignment in the partially semantics-similar pseudo-pair setting. 
We propose Semantics-based Unpaired Multiview Learning (SUM-L), where multiview pseudo-pairs with high similarities are aligned in a semantics-aware manner.
In our method, we also employ video-text alignment to permit all first-person videos to obtain knowledge from samples with different views or modalities.
We highlight that our SUM-L employs the large language model~\cite{text_encoder} to help align cross-view and cross-modality data.
This is different from~\cite{ego-exo}, which proposes to distill beneficial knowledge from unpaired third-person videos~\cite{kinetics-400}, mostly with the help of the off-the-shelf image recognition model~\cite{imagenet} and hand-object detector~\cite{shan2020understanding}.
Moreover, our method is better than existing view-invariant alignment methods, such as typical contrastive learning~\cite{cpc, cmc, dcl, cv_mim} and triplet loss based learning~\cite{face_triplet, yu2019see} in the new setting, since they naively reduce multiview feature distance, which leads to learning suboptimal first-person representations.

To summarize, our contribution is three-fold:
\begin{itemize}
\item We study a new problem: in the partially semantics-similar multiview setting, how to leverage unpaired third-person videos to help first-person video learning.

\item We propose SUM-L for this new problem, which is more effective than existing view-alignment methods in our unpaired multiview learning scenario. 

\item Experiments in standard benchmark datasets including Charades-Ego~\cite{charades-ego}, EPIC-Kitchens~\cite{epic-55}, and EPIC-Kitchens-100~\cite{epic-100} validate the competitive performance of our methods over existing view-alignment methods including typical contrastive learning and triplet loss based learning.
\end{itemize}
\section{Related Work}

\subsection{Egocentric Video Learning}

The study of egocentric video learning has attracted a large volume of attention in recent years.
To verify the quality of model learning, there are multiple egocentric-related tasks, including action recognition~\cite{matsuo2014attention,ego-aco,wang2020symbiotic}, video summarization~\cite{lu2013story,lee2015predicting}, engagement detection~\cite{su2016detecting}, scene affordance~\cite{ego-topo}, gaze prediction~\cite{huang2018predicting}, video question answering~\cite{ego-vqa}, and activity anticipation~\cite{qi2021self,furnari2019would}. 
Nevertheless, aforementioned studies utilize knowledge from first-person videos only~\cite{lu2013story,lee2015predicting,su2016detecting,huang2018predicting,ego-aco,qi2021self,furnari2019would}, or make use of extra supervision from off-the-shelf models~\cite{matsuo2014attention,ego-topo}, or naively pretrain on large-scale datasets~\cite{ego-vqa,wang2020symbiotic}. 
More specifically, these methods do not consider tackling the domain mismatch between different multiview datasets or abstaining from extra supervision. 
Recently, Li~\etal~\cite{ego-exo} try to discover latent signals in the third-person videos which are predictive of key egocentric-specific properties. 
Similar to~\cite{matsuo2014attention,ego-topo}, however, this method heavily relies on the knowledge of off-the-shelf image or video models.

\subsection{First-person Video Representation Learning with Joint Third-person Videos}

The paired first-person and third-person videos, as descriptions of the same actions from different perspectives, are potential to provide beneficial knowledge to each other, yielding more generalizable joint representations.
Recent studies~\cite{charades-ego,charades-ego-plus} propose paired first-person and third-person datasets and learn joint representations of multiview videos.
They show the potential of transferring knowledge from third-person videos to first-person videos. 
In order to extract more useful joint knowledge, Yu~\etal~\cite{yu2020first} make use of the shared attention regions between different views with spatial constraints, in a self-supervised and simultaneous manner. 
However, nowadays very few first-person videos have paired third-person videos~\cite{lemma}.
Moreover, we find that existing multiview methods~\cite{cpc, cmc, face_triplet, yu2019see} learn suboptimal first-person representations from unpaired third-person videos, that are relatively easier to capture and obtain compared with paired data.
Instead, our methodology can better distill unpaired third-person video knowledge to first-person videos,
and it is more efficient and applicable than existing view-alignment methods~\cite{cpc, cmc, face_triplet, yu2019see}.



\subsection{Unpaired Multimodal Learning}

Due to the relatively difficult acquisition of paired multimodal samples and the limited number of them, unpaired multimodal learning recently has attracted the interest of some researchers.
Kundu~\etal~\cite{NEURIPS2021_unpaired_view} treat unpaired 3D pose learning as a self-supervised adaptation problem that aims to transfer the task knowledge from a labeled paired source domain to a completely unpaired target domain.
Next, it proposes the relation distillation method to align the unpaired target videos and unpaired 3D pose sequences.
However, this work makes the assumption that there exist semantically identical cross-modal data in the unpaired target domain, which is almost unavailable in unpaired first- and third-person video representation learning.
Therefore, in this paper, we study a more challenging problem: how to make unpaired multiview alignment in an incompletely semantics-similar situation.
\section{Methodology}


\begin{figure*}[t] 
\centering 
\includegraphics[width=1.0\textwidth]{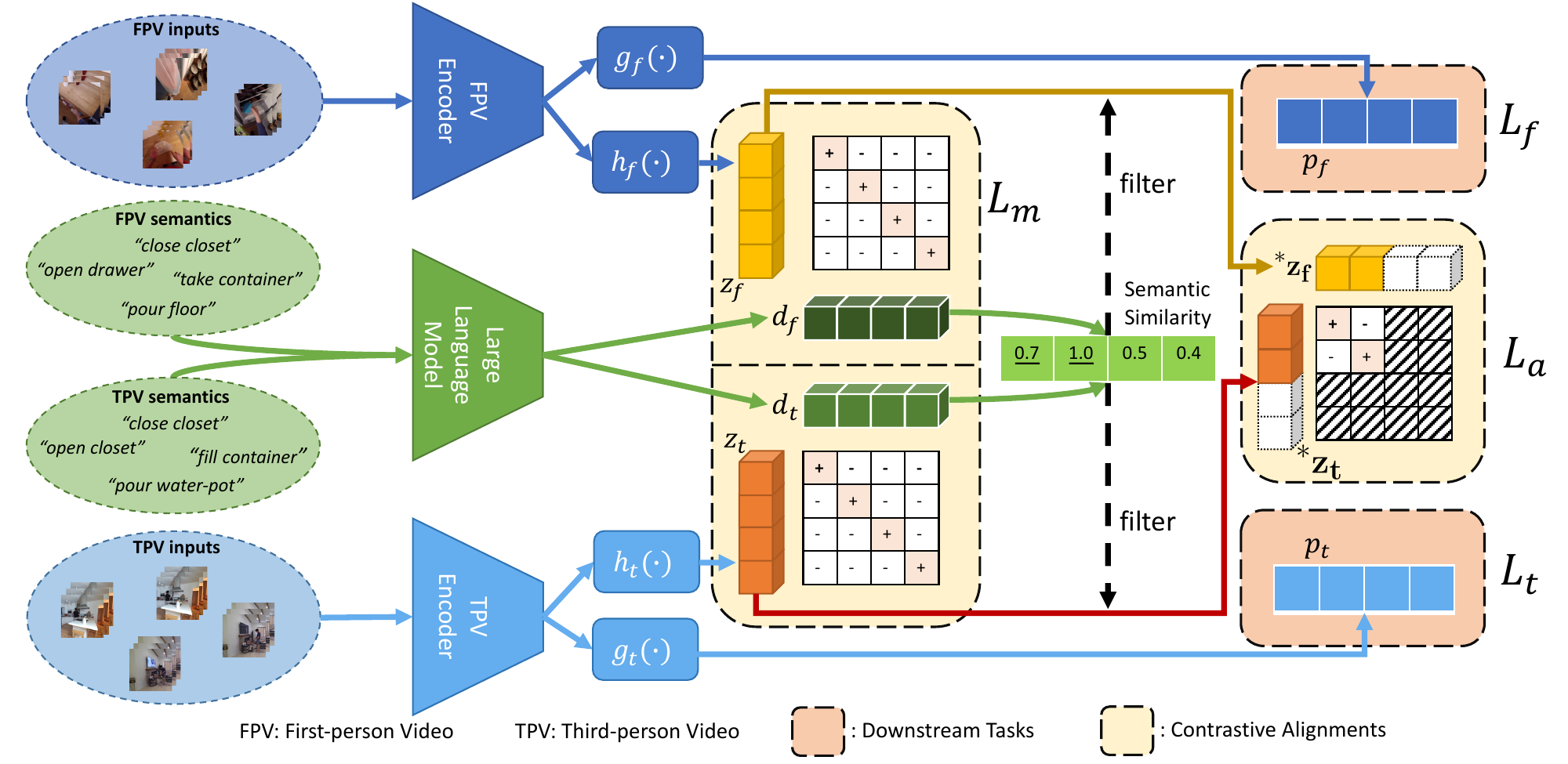} 
\caption{Illustration of our framework. 
First, one batch (batch size $=4$) of multiview pseudo-pairs is built from unpaired first-person and third-person videos. 
The pseudo-pairs are built based on mining the most semantics-similar third-person video for every first-person video. 
During training, the global features for multiview alignment ($z_f, z_t$) are extracted by their corresponding encoders
following the projection networks ($h_f, h_t$). 
In addition, textual features ($d_f, d_t$) are extracted by a large language model~\cite{text_encoder} 
from textual narrations of first-person and third-person videos.
The semantic similarity between $d_f$ and $d_t$ is calculated to filter out the multiview pseudo-pairs with low semantic similarity.
Then the multiview pairs with high semantic similarity are employed to learn the view-invariant representations with the contrastive learning method.
To further improve data efficiency, we employ all the first-person and third-person videos in the batch to learn contrastive multimodal relations.
Finally, the task-specific heads ($g_f, g_t$) for both first-person and third-person videos are trained to make predictions for their corresponding downstream tasks.
During testing, we only use the first-person encoder and task-specific head ($g_f$) to make first-person video predictions.
} 
\label{framework_fig} 
\end{figure*}

In the following sections, we first describe our pseudo-pair construction method (Sec.~\ref{3.1}).
Then we introduce the traditional contrastive framework for paired multiview or multimodal data (Sec.~\ref{3.2}).
Next, we introduce our proposed cross-view and cross-modal methods (Sec.~\ref{3.3}).
Finally, we present our full training and evaluation pipeline in Sec.~\ref{3.4}.

\subsection{Semantics-based Pseudo-pair Mining}
\label{3.1}

Different from the previous joint multiview learning approaches~\cite{charades-ego, charades-ego-plus, yu2019see}, first-person and third-person videos are not automatically paired in the cross-dataset setting.
Hence, before model training, it is essential to construct pseudo multiview pairs.

Every first-person and third-person video has one unique textual narration, which describes the key semantic information of it.
Therefore, we employ the large language model~\cite{text_encoder} to encode textual narrations of multiview videos to vectors.
Then for each first-person video, we mine the third-person video with the highest textual semantic similarity from the third-person video pool to construct the pseudo-pair.
We compute the cosine distance between cross-view video textual narrations as the criterion of semantic similarity between first-person and third-person video clips:
\begin{equation}
    sim_{(\mathbf{z_f}, \mathbf{z_t})} = \frac{\mathbf{d_f} \cdot  \mathbf{d_t}}{\left \| \mathbf{d_f} \right \| \cdot \left \| \mathbf{d_t} \right \|},
\end{equation}
where $\mathbf{d_f}$ and $\mathbf{d_t}$ denote textual narration vectors of first-person and third-person videos, respectively.

\subsection{Typical Multiview Alignment}
\label{3.2}

Existing contrastive-based multiview~\cite{cmc, liu2022view, cv_mim} or multimodal~\cite{albef, tcl, clip, videoclip} methods propose to learn the view-invariant representations via the alignment of the features.
Specifically, they learn to align features between different views or modalities by minimizing the distance between paired samples and maximizing the distance between negative (unpaired) samples.

Given a batch of $N$ normalized first-person features $\{\mathbf{z^1_f}, ..., \mathbf{z^N_f}\}$ and third-person features $\{\mathbf{z^1_t}, ..., \mathbf{z^N_t}\}$, these methods calculate the contrastive multiview Info-NCE~\cite{cmc, homage} loss as:
\begin{equation}
    L_{c} = L^{(f, t)}_{c} + L^{(t, f)}_{c},
\end{equation}
\begin{equation}
\label{info_nce_2}
    L^{(v_1, v_2)}_{c} = -\frac{1}{N}
    \sum_{i=1}^{N}log \left [
    \frac{exp \left ( \left \langle \mathbf{z^i_{v_1}}, \mathbf{z^i_{v_2}} \right \rangle / \tau \right)}
    {exp \left ( \left \langle \mathbf{z^i_{v_1}}, \mathbf{z^i_{v_2}} \right \rangle / \tau \right) + S_i} \right ],
\end{equation}
\begin{equation}
\label{S_i}
    S_i = \sum_{j\in [1, N], j\neq i} 
    exp\left (  \left \langle \mathbf{z^i_{v_1}}, \mathbf{z^j_{v_2}} \right \rangle / \tau \right ),
\end{equation}
where $\mathbf{v_1}$ and $\mathbf{v_2}$ are two views and $\tau$ is the temperature.

For an input video clip $\mathbf{x_{v_k}}$, its normalized feature representation is extracted by the encoder $f_{v_k}(\cdot)$ following the projection network $h_{v_k}(\cdot)$ as:
\begin{equation}
    \mathbf{z_{v_k}} = \frac{h_{v_k}\left (AvgPool\left ( f_{v_k}\left ( \mathbf{x_{v_k}} \right ) \right )  \right )}{\left \| h_{v_k}\left (AvgPool\left ( f_{v_k}\left ( \mathbf{x_{v_k}} \right ) \right )  \right ) \right \|_2}.
\end{equation}

The optimization of contrastive learning methods is often computationally expensive, which requires large batch sizes and a large volume of training epochs.
To tackle this issue, Yeh~\etal~\cite{dcl} refine the InfoNCE loss by removing the positive term from the denominator in Equation~\ref{info_nce_2} and significantly improve the learning efficiency. To be specific, it rewrites Equation~\ref{info_nce_2} as:
\begin{equation}
\label{dc_formula}
\begin{aligned}
    L^{(v_1, v_2)}_{dc} &= -\frac{1}{N}
    \sum_{i=1}^{N}log \left [
    exp \left ( \left \langle \mathbf{z^i_{v_1}}, \mathbf{z^i_{v_2}} \right \rangle / \tau \right) / S_i \right ].
\end{aligned}
\end{equation}

In the context of our unpaired multiview learning, we argue that directly employing the above vanilla contrastive learning method is an obstacle that hinders further acquiring knowledge from third-person videos.
On the one hand, in the unpaired first-person and third-person video multiview representation learning, very few first-person views have the third-person views which share the same semantics.
On the other hand, typical contrastive multiview representation learning is designed to minimize the feature distance between samples with identical semantics, \eg, recordings from camera views for the same objects~\cite{kong2019mmact}, images with different visuals belonging to the same class~\cite{supcon}, one image applying multiple independent~\cite{simclr} or asymmetric~\cite{byol} data augmentations, \etc.
Naively reducing multiview feature distances in our unpaired multiview learning will lead to learning suboptimal first-person video representations.

\subsection{Semantics-based Multiview Learning}
\label{3.3}

To address the above issue, we propose a representation learning method, namely Semantics-based Unpaired Multiview Learning.
We employ contrastive alignment as a tool to align the unpaired multiview videos.
Different from existing view-alignment methods, our proposed method introduces the following two novel alignments:
(1) Semantics-based Cross-view Alignment, which employs the semantic information from textual narrations of multiview videos to align high-similar unpaired data;
(2) Video-text Modal Alignment, which permits all first-person video obtaining knowledge from samples with different views or modalities.
Next, we will introduce our framework step-by-step.


\textbf{Semantics-based Cross-view Alignment.}
We believe that pseudo-pairs with high semantic similarities can better help learn view-invariant information in first-person videos.
Therefore, given cross-view pseudo-pairs $(\mathbf{z_f}, \mathbf{z_t})$ in one batch, we pick pseudo-pairs $(\mathbf{^{*}\textrm{z}_f}, \mathbf{^{*}\textrm{z}_t})$ which have relatively higher textual semantic similarities. 
Then the first-person and third-person video alignment loss $(L_{a})$ is defined as the the decoupled contrastive learning loss~\cite{dcl} of high semantic similarity pseudo-pairs:
\begin{equation}
\label{dcl_formula_ours}
\begin{aligned}
    L_{a}
    =L^{(\mathbf{^{*}\textrm{f}, \mathbf{^{*}\textrm{t}})}}_{dc}
    + L^{(\mathbf{^{*}\textrm{t}}, \mathbf{^{*}\textrm{f}})}_{dc}.
\end{aligned}
\end{equation}

Besides, we hypothesize that the employed large language model~\cite{text_encoder} is a strong teacher: 
by encoding the textual descriptions of videos into vectors, the similarity between vectors implicitly provides a reference to measure the semantic similarity between videos. To make better use of these textual descriptions of first-person and third-person videos, inspired by~\cite{dcl}, we further introduce a semantic-based weighting function for the positive pseudo-pairs.
More specifically, we first rewrite our cross-view video alignment loss $L_{a}$ in Equation~\ref{dcl_formula_ours} as $L_{aw}$, where:
\begin{equation}
    L_{aw} = L^{(\mathbf{^{*}\textrm{f}}, \mathbf{^{*}\textrm{t}})}_{aw} + L^{(\mathbf{^{*}\textrm{t}}, \mathbf{^{*}\textrm{f}})}_{aw}.
\end{equation}
Then given views $(\mathbf{^{*}\textrm{f}}, \mathbf{^{*}\textrm{t}})$, we have:
\begin{equation}
\label{dclw_2}
    L^{(\mathbf{^{*}\textrm{f}}, \mathbf{^{*}\textrm{t}})}_{aw}
    =\frac{1}{N} \sum_{i=1}^{N}
    \left [ log \left ( S_i \right ) - \omega \left ( \mathbf{^{*}\textrm{d}^{i}_{f}}, \mathbf{^{*}\textrm{d}^{i}_{t}} \right )\cdot 
    \left \langle \mathbf{^{*}\textrm{z}^{i}_{f}}, \mathbf{^{*}\textrm{z}^i_{t}} \right \rangle / \tau \right ],
\end{equation}
where the definition of $S_i$ is the same as Equation~\ref{S_i}.
Here we implement the weighting function as:
\begin{equation}
    \omega \left ( \mathbf{^{*}\textrm{d}^{i}_{f}}, \mathbf{^{*}\textrm{d}^{i}_{t}} \right ) = 
    \frac{exp\left(\left \langle \mathbf{^{*}\textrm{d}^{i}_{f}}, \mathbf{^{*}\textrm{d}^{i}_{t}} / \sigma  \right \rangle \right) }
    {\mathbb{E}_{i} \left[ exp\left(\left \langle \mathbf{^{*}\textrm{d}^{i}_{f}}, \mathbf{^{*}\textrm{d}^{i}_{t}} / \sigma  \right \rangle \right) \right]},
\end{equation}
where the higher textual similarity pseudo-pair has, the more learning signal the pseudo-pair needs to be provided, and vice versa.
In this case, our multiview alignment is semantics-aware, which is different from existing multiview alignment methods~\cite{cmc, yu2019see} that evenly align multiview data with different semantic similarities.

\textbf{Video-text Modal Alignment.} Until now, although the above statement provides a seemingly finalized solution in the case where the multiview pseudo-pair semantic information is not exactly the same, it is not data-efficient for every first-person sample.
This is because it cannot help first-person videos in pseudo-pairs with low semantic similarities gain learning signals from others.

In order to permit all first-person videos to obtain knowledge from samples with different views or modalities, we employ video-text contrastive alignment to make the model training more data-efficient.
As we mentioned before that every first-person and third-person video has an accurate textual description of their semantic information, this multimodal alignment can further help the neural network learn useful knowledge from these phrases.

Technically, given a batch of $N$ normalized first-person and third-person features $\{\mathbf{z^1_f}, ..., \mathbf{z^N_f}\}$ and $\{\mathbf{z^1_t}, ..., \mathbf{z^N_f}\}$, and their corresponding normalized textual description vectors $\{\mathbf{d^1_f}, ..., \mathbf{d^N_f}\}$ and $\{\mathbf{d^1_t}, ..., \mathbf{d^N_t}\}$, the multimodal contrastive alignment loss $(L_{m})$ for both first-person videos and third-person videos is defined as:
\begin{equation}
\label{formula_mm}
\begin{aligned}
    L_{m} = 
    L^{(\mathbf{z_{f}},\mathbf{d_{f}})}_{dc} + 
    L^{(\mathbf{d_{f}},\mathbf{z_{f}})}_{dc} + 
    L^{(\mathbf{z_{t}},\mathbf{d_{t}})}_{dc} + 
    L^{(\mathbf{d_{t}},\mathbf{z_{t}})}_{dc}
\end{aligned}.
\end{equation}

\subsection{Training and Evaluation Pipelines}
\label{3.4}

The two proposed cross-view and cross-modal losses, along with the task-specific losses for both first-person videos $(L_f)$ and third-person videos $(L_t)$, are combined together during the training procedure to construct the final training loss:
\begin{equation}
\label{overall_loss}
    \mathbb{L} = L_f + w_t \cdot L_{t} + w_{aw} \cdot L_{aw} + w_{m} \cdot L_{m}.
\end{equation}

Note that our model is trained in multiple stages. First, both first-person and third-person video encoders are initialized by weights pretrained on Kinetics-400~\cite{kinetics-400}.
Then we train the third-person video encoder and task-specific head on the third-person video dataset with third-person video task-specific loss $(L_t)$ only.
Next, we load this trained weight for the third-person video encoder and task-specific head and do training for the whole framework with the overall loss in Equation~\ref{overall_loss}.
Finally, we evaluate the first-person video performance of our method using the first-person video encoder and task-specific head.

The specific task for both first-person and third-person videos we discussed in this paper is video classification.
Therefore, both $L_f$ and $L_t$ are classification losses; $w_t$, $w_{aw}$, and $w_m$ are the loss weights. 
\section{Experiments}

\subsection{Experiment Settings}

\subsubsection{Dataset}

In this section, we introduce the standard benchmarks we used for evaluation in the experiments.


\textbf{LEMMA}~\cite{lemma} contains 648 third-person videos and 445 first-person videos. 
Among them, 136 activities were performed in kitchens, and the remaining 188 were in the living rooms.
It contains 25 verb classes, 64 noun classes, and 863 action classes in their action recognition task, and includes not only single-agent but also multi-agent videos. 
We utilize single-agent third-person videos only as our unpaired third-person samples.

\textbf{Charades-Ego}~\cite{charades-ego,charades-ego-plus} contains 68,536 activity instances in paired 68.8 hours of first-person and third-person videos.
It includes 157 action classes. 
We only utilize first-person videos during training and inference.

\textbf{EPIC-Kitchens}~\cite{epic-55} is a large-scale egocentric video benchmark recorded in kitchen environments. 
It contains 55 hours of video with a total of 39.6K action segments and is labeled with 352 object and 125 verb classes.

\textbf{EPIC-Kitchens-100}~\cite{epic-100} extends EPIC-Kitchens~\cite{epic-55} to 100 hours and 89,977 action segments. 
We employ this as well as EPIC-Kitchens~\cite{epic-55} to validate our approach.

\subsubsection{Baseline Methods}

We compare the proposed approach with the following baseline methods, including typical multiview representation learning methods:

\textbf{FPV Only} is trained with first-person (FPV) video data only. Its framework is composed of first-person video encoder and first-person video classification head. 

\textbf{Typical CL} is widely employed for view- or modal-invariant contrastive learning (CL)~\cite{cmc, cv_mim}, and vision-language models~\cite{clip, videoclip, albef, tcl, egovlp, Zhao2022LearningVR}. 
Here we implement it with the loss function in Equation~\ref{dc_formula}. 
Similar to our framework, videos from different views are fed to different encoders.
We employ the same pseudo-pair construction method and the same inference strategy as our framework for this baseline method.

\textbf{Triplet} is another widely used multiview representation learning method~\cite{yu2019see, charades-ego}.
We employ the same implementation protocol following~\cite{yu2019see}. 
Moreover, we leverage this method to tackle our problem following the same  procedures as ``Typical CL''.

\subsubsection{Implementation Details}

To ensure fairness, we keep all hyperparameters the exact same for all baselines and our method.
For Charades-Ego~\cite{charades-ego,charades-ego-plus}, we use the base learning rate of 0.25, batch size of 32 and maximum epoch of 60.
For EPIC-Kitchens~\cite{epic-55} and EPIC-Kitchens-100~\cite{epic-100}, we use the base learning rate of 0.01, weight decay of 0.0001, batch size of 64 and maximum epoch of 30.
Following~\cite{ego-exo}, all models are trained using the stochastic gradient descent optimizer~\cite{sgd} with the momentum of 0.9.
For Charades-Ego~\cite{charades-ego,charades-ego-plus}, we use the cosine learning rate decay.
For EPIC-Kitchens~\cite{epic-55} and EPIC-Kitchens-100~\cite{epic-100}, we use sequential learning rate decay.
The backbones we use are ResNet-3D-50~\cite{r3d} and SlowFast~\cite{slowfast}.
More details about the implementation of backbones will be provided in the following experimental sections.

\subsection{Validity of Proposed Method}


Table~\ref{bs_com} shows the performance of different training strategies, where all methods are implemented with the ResNet-3D-50~\cite{r3d} encoder.
Our proposed method outperforms the ``FPV only'' baseline model on both Charades-Ego and EPIC-Kitchens
(+1.6\% mAP on Charades-Ego, and +1.2\%/+1.5\% on EPIC-Kitchens verbs/nouns top-1 accuracies). 
This demonstrates that under partially similar semantics situations, with our method, the third-person videos can help to improve first-person video representations. 

Moreover, different from our method, previous multiview learning methods (``Typical CL'' and ``Triplet'') cannot bring significant and consistent improvement compared with ``FPV only''. 
For example, ``Typical CL'' brings only 0.4\% improvement on Charades-Ego (far less than ours) and even deteriorates performance on EPIC-Kitchens.
Instead, our proposed method outperforms existing alignment methods up to +1.2\% mAP on Charades-Ego and +1.8\%/+2.0\% on EPIC-Kitchens verbs/nouns top-1 accuracies.
We believe this is due to the following reasons.
(1) Our method semantics-awarely aligns unpaired data, instead of evenly aligning pseudo-pairs with different semantic similarities.
(2) Our method filters the negative effects caused by multiview pseudo-pairs of low semantic similarity.
This is different from typical multiview methods which naively align all the pairs.
(3) Our method permits all first-person videos to obtain knowledge from samples with different views or
modalities, which improves data efficiency.

\begin{table}[t]
\centering
\resizebox{\linewidth}{!}{
\caption{Comparison with baseline methods. 
The backbone we use is R3D-50~\cite{r3d} for all methods. 
``C-Ego'' denotes Charades-Ego~\cite{charades-ego} dataset and ``EK-55'' denotes EPIC-Kitchens~\cite{epic-55} dataset.}
\setlength{\tabcolsep}{1mm}{
\begin{tabular}{c|c|cc|cc}
\hline
Dataset     & C-Ego & \multicolumn{2}{c|}{EK-55 (Verb)} & \multicolumn{2}{c}{EK-55 (Noun)} \\ \hline
Evaluations & mAP (\%)  & \begin{tabular}[c]{@{}c@{}}Top-1 \\ acc. (\%)\end{tabular} & \begin{tabular}[c]{@{}c@{}}Top-5 \\ acc. (\%)\end{tabular} & \begin{tabular}[c]{@{}c@{}}Top-1 \\ acc. (\%)\end{tabular} & \begin{tabular}[c]{@{}c@{}}Top-5 \\ acc. (\%)\end{tabular} \\ \hline
FPV Only    & 24.7          & 61.1          & 87.1          & 45.8          & 70.4  \\
Typical CL  & 25.1          & 60.5          & 86.2          & 45.3          & 68.6  \\
Triplet     & 25.2          & 61.5          & 86.8          & 45.7          & 69.7  \\ \hline
SUM-L (Ours)        & \textbf{26.3} & \textbf{62.3} & \textbf{87.2} & \textbf{47.3} & \textbf{70.9} \\ \hline
\end{tabular}}
\label{bs_com}
}
\end{table}

We further verify the effectiveness of our proposed method with a larger backbone, SlowFast~\cite{slowfast}, as the encoder, and we compare with recent state-of-the-art methods on three popular egocentric action recognition benchmarks, Charades-Ego~\cite{charades-ego}, EPIC-Kitchens~\cite{epic-55}, and EPIC-Kitchens-100~\cite{epic-100}.
Note that evaluation scores of state-of-the-arts are reported directly from their paper.
The experimental results employing SlowFast~\cite{slowfast} encoder reconfirm the following findings.
(1) Our method is effective for cross-view pseudo-pair alignment when there is only partially same multiview semantics. 
(2) Compared with existing multiview alignment methods, the improvement based on ``FPV Only'' of our method is still more significant and consistent.
Next, we analyze the important details of the experiments.

\begin{table}[t]
\small
\centering
\caption{Comparison on Charades-Ego. Our method and all baseline methods are implemented with SlowFast-R101~\cite{slowfast} backbone. ``EE'' denotes training based on pretrained weights provided by~\cite{ego-exo}.}
\begin{tabular}{c|c|c}
\hline
                                   & Methodology & mAP (\%)      \\ \hline
\multirow{5}{*}{state-of-the-arts} & Actor~\cite{charades-ego}       & 20.0 \\
                                   & SSDA~\cite{ssda}        & 23.1 \\
                                   & I3D~\cite{ego-exo}         & 25.8          \\
                                   & Ego-Exo~\cite{ego-exo}     & 28.7          \\
                                   & Ego-Exo*~\cite{ego-exo}    & 30.1          \\ \hline
\multirow{4}{*}{w/o ``EE''}          & FPV Only    & 27.4          \\
                                   & Typical CL  & 27.9          \\
                                   & Triplet     & 28.1          \\
                                   & SUM-L (Ours)        & 28.5          \\ \hline
\multirow{4}{*}{w/ ``EE''}           & FPV Only    & 29.5          \\
                                   & Typical CL  & 29.1          \\
                                   & Triplet     & 28.6          \\
                                   & SUM-L (Ours)        & \textbf{30.7}          \\ \hline
\end{tabular}
\label{sota_charade}
\end{table}

\textbf{Comparison on Charades-Ego} Table~\ref{sota_charade} demonstrates the effectiveness of our method for cross-view pseudo-pair alignment when there is only partially same semantics.
Note that our method and Ego-Exo~\cite{ego-exo} complement each other. 
Thus, we can also incorporate our method into Ego-Exo~\cite{ego-exo}. 
As shown in Table~\ref{sota_charade}, based on pretrained weight on~\cite{ego-exo}, our proposed method further improves the performance of ``FPV only'' by 1.2\% mAP, and achieves a new state-of-the-art on Charades-Ego~\cite{charades-ego}.

\begin{table}[t]
\centering
\resizebox{\linewidth}{!}{
\caption{Comparison on EPIC-Kitchens. The accuracies are measured by top-1 verb \& noun classification accuracies. Our method and all baseline methods are implemented with SlowFast-R50~\cite{slowfast} backbone.}
\begin{tabular}{c|c|cc}
\hline
                                   & Methodology  & Verb (\%)     & Noun (\%) \\ \hline
\multirow{3}{*}{state-of-the-arts} & SlowFast-R50~\cite{slowfast} & 64.1 & 48.6      \\
                                   & Ego-Exo~\cite{ego-exo}      & 66.0 & 49.4      \\
                                   & Ego-Exo*~\cite{ego-exo}      & 66.4          & 49.8      \\ \hline 
\multirow{4}{*}{w/o ``EE''}          & FPV Only     & 63.1          & 48.4      \\
                                   & Typical CL   & 63.9          & 49.6         \\
                                   & Triplet      & 63.9          & 48.8      \\
                                   & SUM-L (Ours)         & 65.0          & 49.5      \\ \hline
\multirow{4}{*}{w/ ``EE''}           & FPV Only     & 65.8          & 49.5      \\
                                   & Typical CL   & 63.8          & 48.4      \\
                                   & Triplet      & 63.9          & 49.6      \\
                                   & SUM-L (Ours)         & \textbf{66.7}    & \textbf{50.2}      \\ \hline 
\end{tabular}
\label{sota_ep55}}
\end{table}

\textbf{Comparison on EPIC-Kitchens} As we can see from Table~\ref{sota_ep55}, our method still brings clear improvement compared with ``FPV Only'' method (+1.9\%/+1.1\% verbs/nouns top-1 accuracies).
Furthermore, with the help of pretrained weights from~\cite{ego-exo}, our method still outperforms most state-of-the-art methodologies, which indicates the effectiveness of our method.
Last but not least, our method outperforms existing alignment methods up to +2.9\%/+1.8\% of verbs/nouns top-1 accuracies.

\begin{table}[t]
\footnotesize
\centering
\resizebox{\linewidth}{!}{
\caption{Comparison on EPIC-Kitchens-100 with state-of-the-arts that implement CNN backbone. The accuracies are measured by top-1 verb \& noun classification accuracies. We implement SlowFast-R101~\cite{slowfast} backbone.}
\begin{tabular}{c|c|cc}
\hline
                                   & Methodology & Verb (\%)     & Noun (\%) \\ \hline
\multirow{6}{*}{state-of-the-arts} & TSN~\cite{tsn}         & 60.2 & 46.0      \\
                                   & TSM~\cite{tsm}         & 67.9 & 49.0      \\
                                   & TempAgg~\cite{tempagg}     & 59.9          & 45.1      \\
                                   & Ego-Exo~\cite{ego-exo}     & 67.0          & 52.9      \\
                                   & Ego-Exo*~\cite{ego-exo}    & 67.5          & 52.9      \\
                                   & IPL~\cite{ipl}         & \textbf{68.6}          & 51.2      \\ \hline
w/ ``EE''                             & SUM-L (Ours)        & 67.0          & \textbf{53.4}       \\ \hline
\end{tabular}
\label{sota_ep100}}
\end{table}

\textbf{Comparison on EPIC-Kitchens-100}  
As shown in table~\ref{sota_ep100}, our method also obtains competitive results on EPIC-Kitchens-100~\cite{epic-100} dataset when trained based on weights from~\cite{ego-exo}. 
It is worth noting that our method gains the first place in noun accuracy and third place in verb accuracy compared with state-of-the-art methodologies which employ convolutional neural network backbones.

\subsection{Ablation Studies}

In this section, we conduct the ablation studies on Charades-Ego~\cite{charades-ego} dataset.
The backbone we employ is ResNet-3D-50~\cite{r3d}.

\subsubsection{Impact of Different Alignment Tasks}

The experiment results are shown in Table~\ref{ab_tasks}.
We find our proposed unpaired cross-view alignment method itself brings 0.9\% mAP improvement comparing with the ``FPV Only'' method.
It denotes that the method can help learn useful knowledge from unpaired third-person view samples when the cross-view semantic information is not identical.
Moreover, our proposed video-text modal alignment method further boosts first-person performance by 0.7\% mAP compared with implementation with cross-view alignment only.
We think this is because the proposed video-text modal alignment ensures the alignment of all first-person video samples, and further improves the semantics utilization for both first-person and third-person videos.
Besides, implementing our cross-view alignment without semantic weighting causes a slight 0.2\% drop, which manifests the usefulness of knowledge from the large language model.
This knowledge provides a reference to measure the semantic similarity between cross-view videos.

\begin{table}[t]
\centering
\caption{
Different training task combinations of our proposed method. 
Combining semantics-aware unpaired cross-view alignment with video-text alignment, our method clearly improves egocentric classification performance on the Charades-Ego dataset. }
\setlength{\tabcolsep}{1mm}{
\begin{tabular}{ccc|c}
\hline
$L_{a}$ & $L_{aw}$ & $L_{m}$ & mAP (\%)   \\ \hline
$\times$         & $\times$          & $\times$        & 24.7      \\ \hline
$\times$         & $\checkmark$          & $\times$        & 25.6          \\
$\times$         & $\checkmark$          & $\checkmark$        & 26.3      \\ 
$\checkmark$         & $\times$          & $\checkmark$        & 26.1 \\ \hline
\end{tabular}}
\label{ab_tasks}
\end{table}

\begin{figure*}[t] 
\centering 
\includegraphics[width=0.96\textwidth]{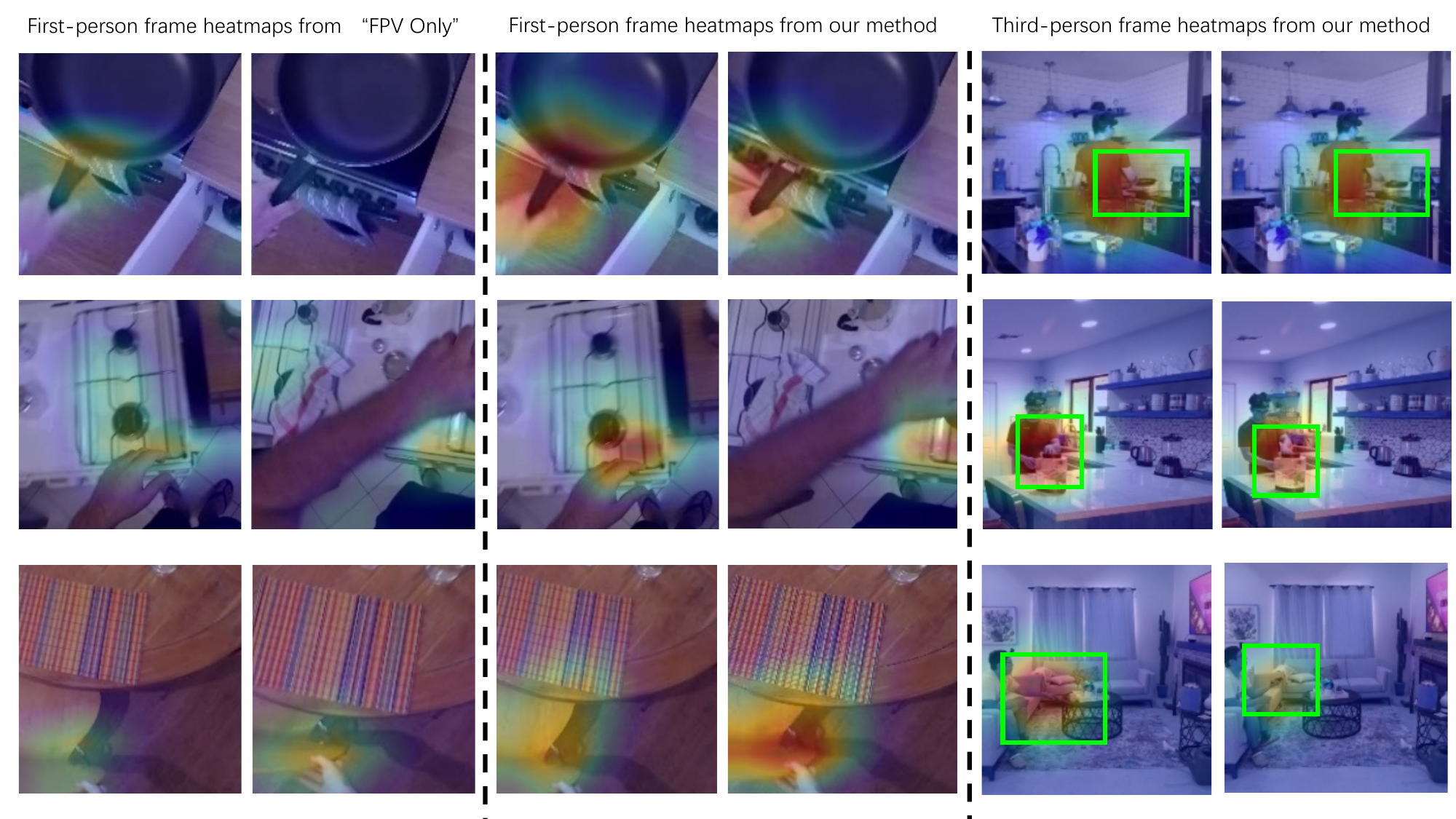} 
\caption{Comparison of class activation map visualizations between the ``FPV Only'' method and our method.
The cross-view similar semantics for data in the first row is ``holding pan''. 
The cross-view similar semantics for data in the second and third rows is ``take''.
} 
\label{vis_fig} 
\end{figure*}

\subsubsection{Impact of Third-person View Backbone}

The original setup of the third-person video encoder is mentioned in Sec.~\ref{3.4}.
Here we study the effect of other intuitively feasible third-person video encoder implementations.
The experiment results are shown in Table~\ref{ab_tpv}.

First, we load the pretrained weight of the third-person video and freeze it during the training process for the whole framework shown in Figure~\ref{framework_fig}.
We find freezing the third-person video encoder causes a 0.5\% mAP drop compared with the original setup. 
This indicates that for the pretrained third-person video encoder, allowing training more parameters can better help first-person video learn effective information from third-person samples.
Moreover, naively employing shared encoder weights for different views, which follows~\cite{simclr}, can cause clear performance drop (1.0\% mAP drop) to first-person video learning.
We think this is due to the big variances between first-person and third-person samples.
Finally, the same initialization (pretrained on Kinetics-400~\cite{kinetics-400}) for both first-person and third-person video encoders causes a slight 0.2\% mAP drop.
It indicates that for the third-person video encoder, employing existing weights of third-person samples for optimization has a positive effect on first-person video learning.

\begin{table}[t]
\centering
\caption{Impact of different third-person video encoder implementations. The ``Original'' setup is described in Sec.~\ref{3.4}.}
\begin{tabular}{c|c}
\hline
Setup                                & mAP (\%) \\ \hline
Freezed TPV Backbone                 & 25.8    \\
Shared Multiview Encoder Weights     & 25.3    \\
Same Initialization with FPV Encoder & 26.1    \\ \hline
Original                             & 26.3    \\ \hline
\end{tabular}
\label{ab_tpv}
\end{table}

\subsection{Visualization}

We visualize the first-person and third-person video class activation maps (CAM)~\cite{cam} on EPIC-Kitchens~\cite{epic-55} dataset.
We employ first-person video encoder weights from ``FPV Only'' method (61.1\% \& 45.8\% top-1 verb \& noun accuracies) and our method (62.3\% \& 47.3\% top-1 verb \& noun accuracies).
As shown in the first and second columns of Figure~\ref{vis_fig}, comparing with ``FPV Only'' method, our method can help the model be more responsive to positive spatio-temporal locations of video clips.

Moreover, from the third column of Figure~\ref{vis_fig}, we find that our model also has a strong response to the third-person video locations which have overlapped semantics with the first-person video (see bright green bounding boxes of Figure~\ref{vis_fig}).
It shows that our method successfully utilizes knowledge from unpaired third-person videos to help learn first-person video representations.
\section{Conclusions}

In this paper, we tackle a new problem in first-person and third-person video learning and propose SUM-L.
It can distill knowledge from unpaired, cross-dataset third-person video without exact same multiview semantic information. 
Our experiments show the validity of our proposed method.
Besides, under the more challenging scenario than typical paired or unpaired multimodal or multiview learning, our method outperforms existing view-alignment methods.

However, improvements from SUM-L are not very significant.
One possible reason is due to the relatively small amount of third-person data.
It will be interesting to explore more unpaired third-person video datasets.
It will be also interesting to study more downstream tasks, such as unpaired cross-view retrieval.
We leave these as future work.

\noindent\textbf{Acknowledgements} 
This work is partially supported by National Science Foundation (NSF) CMMI-2039857, and University of Delaware Research Foundation (UDRF). 
We thank Florian Schroff and Hartwig Adam from Google Research for supporting this project.

{\small
\bibliographystyle{ieee_fullname}
\bibliography{egbib}
}

\end{document}